\documentclass[a4paper]{article}
\usepackage[numbers,sort&compress]{natbib}
\usepackage{INTERSPEECH2020}
\usepackage{multirow}
\usepackage{hyperref}

\title{Voting for the right answer: Adversarial defense for speaker verification}
\name{Haibin Wu$^{1,2*}$, Yang Zhang$^{1*}$, Zhiyong Wu$^{1\dagger}$, Dong Wang$^{3\dagger}$, Hung-yi Lee$^{2\dagger}$
\thanks{* Equal contribution.}
\thanks{$\dagger$ Equal correspondence.}
}

\address{
  $^1$Tsinghua-CUHK Joint Research Center for Media Sciences, Technologies and Systems,\\
    Shenzhen International Graduate School, Tsinghua University, Shenzhen\\
  $^2$Graduate Institute of Communication Engineering, National Taiwan University, Taipei\\
  $^3$Center for Speech and Language Technologies, Tsinghua University, Beijing
  }

\begin{document}

\maketitle
\begin{abstract}
Automatic speaker verification (ASV) is a well developed technology for biometric identification, and has been ubiquitous implemented in security-critic applications, such as banking and access control.
However, previous works have shown that ASV is under the radar of adversarial attacks, which are very similar to their original counterparts from human's perception, yet will manipulate the ASV render wrong prediction.
Due to the very late emergence of adversarial attacks for ASV, effective countermeasures against them are limited.
Given that the security of ASV is of high priority, in this work, we propose the idea of ``voting for the right answer" to prevent risky decisions of ASV in blind spot areas, by employing random sampling and voting.
Experimental results show that our proposed method improves the robustness against both the limited-knowledge attackers by pulling the adversarial samples out of the blind spots, and the sufficient-knowledge attackers by introducing randomness and increasing the attackers' budgets.

\end{abstract}


\section{Introduction}
\label{sec:intro}
Automatic speaker verification (ASV), is to determine whether a claimed identity is speaking during a speech segment \cite{zheng2017robustness}.
Thanks to previous efforts \cite{kenny2007joint,dehak2010front,li2017deep,snyder2018x}, ASV is now a well-developed technology for biometric identification and widely adopted in various security-critic applications.
But, ASV models with high performance are vulnerable to spoofing audios \cite{todisco2019asvspoof} generated by audio replay, text-to-speech and voice conversion, back-door attacks \cite{zhai2021backdoor}, and related emerged adversarial attacks \cite{kreuk2018fooling,das2020attacker}.
In this paper, we mainly focus on tackling the adversarial attacks.

Adversarial samples were first raised by \cite{szegedy2013intriguing} and they regarded them as blind spots of models.
Harnessing adversarial samples to attack well-trained models is \emph{adversarial attack}.
Adversarial samples are generated by slightly modifying the genuine samples with deliberately crafted perturbations, and \cite{szegedy2013intriguing} shows that such elaborated samples can manipulate image classification models with high performance predict wrong answers.
Also, audio processing models are vulnerable to adversarial attacks.
Carlini and Wagner \cite{carlini2018audio} shows that adversarial attacks can hallucinate state-of-the-art automatic speech recognition (ASR) model.
Given a piece of audio, either speech, silence, or music, \cite{carlini2018audio} can generate another perceptually indistinguishable adversarial sample, which can deceive the ASR output text predefined by the attackers.
The vulnerability of ASR to adversarial attacks has been also studied by \cite{yuan2018commandersong,schonherr2018adversarial,yakura2018robust,qin2019imperceptible}.
Other audio processing tasks, such as anti-spoofing for ASV \cite{liu2019adversarial,wu2020defense_2,wu2020defense}, voice conversion \cite{huang2021defending} and sound event classification \cite{subramanian2019robustness}, are also subject to adversarial attacks.

The performance of ASV will drop catastrophically under adversarial attacks \cite{kreuk2018fooling,li2020adversarial,xie2020real,marras2019adversarial,li2020practical,chen2019real,wang2020inaudible,villalba2020x,das2020attacker}.
\cite{kreuk2018fooling} firstly shows ASV is subject to adversarial attacks.
\cite{chen2019real} attacks ASV trained by only a few speakers.
\cite{villalba2020x} and \cite{li2020adversarial} respectively show the vulnerability to adversarial attacks of i-vector and x-vector models.
Recently, researchers also investigated more risky adversarial attacks which are inaudible \cite{wang2020inaudible} and universal \cite{marras2019adversarial,xie2020real}, and can spread over the air \cite{li2020practical,xie2020real}.
Only limited works \cite{wang2019adversarial,li2020investigating,zhang2020adversarial,wu2021adversarialasv} are conducted to protect ASV from adversarial attacks.
So it is still an open question to effectively tackle this issue.
\cite{zhang2020adversarial} trains a deep neural network based filter to eliminate adversarial noise.
A binary classification model \cite{li2020investigating} is trained to spot the adversarial samples.
\cite{wang2019adversarial} adopts adversarial training  to improve the robustness of ASV.
Self-supervised learning based filters are used to alleviate the adversarial noise for ASV \cite{wu2021adversarialasv,wu2021adversarial-2}.

In this paper, we propose to improve the adversarial robustness by sampling the neighbors of a given utterance and letting the neighbors vote for whether the utterance should be accepted or not by the ASV model.
The adversarial samples are blind spots of ASV and we believe the radius of such blind spots is small, as shown in Table~\ref{tab:defense_results}.
So if the spread of the sample space is large, then there is a large probability that most of the neighbours are outside of the blind spot. 
In this case, the voting will result in the right decision.
Contrary to \cite{wang2019adversarial,li2020investigating,zhang2020adversarial} which require the system designers to model the in-the-wild attackers and know exactly the adversarial attack methods adopted during attack in advance, our method is attack-agnostic.
In contrast to \cite{wu2021adversarialasv,wu2021adversarial-2} which counter adversarial noise in the frequency domain, our method is in the time domain.
Our method harnesses the idea of ``voting for the right answer" to let the ASV make the decision based on an utterance's neighbors rather than the utterance itself, thus equips the ASV model with the capacity of countering adversarial attacks.
The experimental results show the effectiveness of our defense method.

\section{Background}

\subsection{Automatic Speaker verification}
ASV aims to determine whether a piece of given speech belongs to an alleged speaker.
Given an enrollment and test utterance, $x_{e}$ and $x_{t}$, the ASV model will certify whether they are pronounced by the same speaker.
Specifically, the procedure of ASV can be divided into three steps: firstly feature engineering which maps the audio waveform to acoustic features, secondly speaker embedding extraction which extracts the fixed-dimensional speaker embeddings from the variable-length acoustic feature sequences, and finally a back-end model that measures the similarity between speaker embeddings.
For brevity, the three steps can be denoted by a scoring function $f$:
\begin{align}
    &s = f(x_{t}, x_{e}), \label{eq:score}
\end{align}
where, $x_{t}$ and $x_{e}$ are the test and enroll utterance, respectively; $s$ measures the similarity between $x_t$ and $x_e$.
The smaller $s$ is, the less likely they belong to the same person, and vice versa.

\subsection{Adversarial attack}
Adversarial samples are generated by slightly modifying the genuine samples with deliberately crafted perturbations.
The deliberately crafted perturbations are usually indistinguishable from human's perception, yet significantly change the output of the model.
Different strategies to generate the adversarial noise result in different attack algorithms, and in this paper, we adopt Basic Iterative Method (BIM) \cite{kurakin2016adversarial}, an efficient iterative adversarial attack method.
In BIM, attackers initialize $x_{t}^{0}=x_{t}$, where $x_{t}$ means the testing utterance in Eq.~\ref{eq:score},then iteratively update it as this equation:
\begin{equation}
\begin{aligned}
    x_{t}^{n+1}=Clip_{\epsilon}^{x_{t}}\left(x_{t}^{n}+\alpha \cdot \lambda \cdot sign\left(\nabla_{x_{t}^{n}}f(x_{t}^{n}, x_{e}) \right)\right), 
    \\ for \, n=0,1, \ldots, N-1,
\end{aligned}
\end{equation}
where $Clip(.)$ is a clipping  operation that ensures $||x_{t}^{n+1} - x_{t}||_{\infty}\leq \epsilon$, $\epsilon \geq 0$ and $\epsilon \in \mathbb{R}$, $\alpha$ is the step size, $\lambda=1$ for the non-target trial and $\lambda=-1$ for the target trial, and $N$ is the number of iterations.
$\epsilon$ determines the attack intensity which is a predefined value by attackers.
In the target trial, the testing and enroll speech are uttered by the same speaker, while in the non-target trial, they are generated by different speakers.
For example, if the attacker want to attack ASV for the non-target trial, they will adopt BIM to make the similarity score between the testing and enrollment speech as high as possible in order to let the ASV model falsely accept the imposter.

\subsection{System evaluation metrics}
The testing trials can be represented as: $\mathbb{T}=\mathbb{T}_{tgt} \cup \mathbb{T}_{ntgt}$, where $\mathbb{T}_{tgt}$ and $\mathbb{T}_{ntgt}$ denote the target trials and non-target trials respectively.
We divide the testing trials into two sets, $\mathbb{T}=\mathbb{D} \cup \mathbb{E}$, where $\mathbb{D}=\mathbb{D}_{tgt} \cup \mathbb{D}_{ntgt}$ is the development set, $\mathbb{E}=\mathbb{E}_{tgt} \cup \mathbb{E}_{ntgt}$ is the evaluation set.
We determine the decision threshold $\tau_{ASV}$ by the development set as below:
\begin{align}
    &DevFAR(\tau) = \frac{\vert \{s_i \geq \tau : i \in \mathbb{D}_{ntgt}\} \vert}{\vert \mathbb{D}_{ntgt} \vert} \label{eq:gen-far} \\
    &DevFRR(\tau) = \frac{\vert \{s_i < \tau : i \in \mathbb{D}_{tgt}\} \vert}{\vert \mathbb{D}_{tgt} \vert} \label{eq:gen-frr} \\
    &\tau_{ASV} = \{ \tau \in \mathbb{R} : DevFAR(\tau) = DevFRR(\tau) \} \label{eq:asv-threshold} 
\end{align}
where $DevFAR(\tau)$ and $DevFRR(\tau)$ mean the development false acceptance rate and false rejection rate respectively under the decision threshold $\tau$, $\mathbb{D}_{tgt}$ and $\mathbb{D}_{ntgt}$ denote the development trial sets containing target and non-target trials, respectively, $s_i$ is the ASV score for the trial $i$ and $\vert \mathbb{A} \vert$ denotes the number of elements in set $\mathbb{A}$. 

Then we evaluate the ASV performance on evaluation set, by FAR and FRR:
\begin{align}
    &FAR = \frac{\vert \{s_i \geq \tau_{ASV} : i \in \mathbb{E}_{ntgt} \} \vert}{\vert \mathbb{E}_{ntgt} \vert} \label{eq:adv-far} \\
    &FRR = \frac{\vert \{ s_i < \tau_{ASV} : i \in \mathbb{E}_{tgt} \} \vert}{\vert \mathbb{E}_{tgt} \vert} \label{eq:adv-frr}
\end{align}
where $\tau_{ASV}$ is the decision threshold determined by Eq.~\ref{eq:asv-threshold}.

\section{Voting for the right answer}

\begin{figure}[ht]
  \centering
  \centerline{\includegraphics[width=7cm]{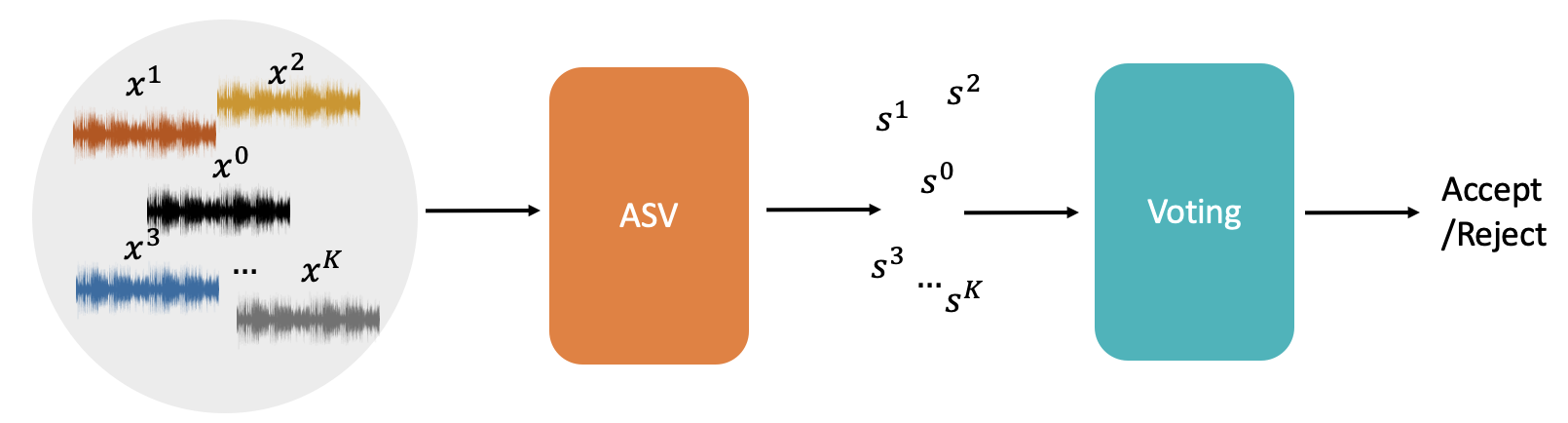}}
  \caption{Proposed framework.}
  \label{fig:method}
\end{figure}

\subsection{Proposed method}

In this subsection, we only introduce the procedure of our proposed method, and detailed explanation and intuition of why it works will be illustrated in next subsection.
For the sake of simplicity, we omit the enroll utterance $x_{e}$, and denote the testing utterance $x_{t}$ as $x^{0}$.
As shown in Fig.~\ref{fig:method}, a Gaussian ball with $x^{0}$ as the center, and variance ,$\sigma$, as the radius, is constructed.
Then we sample $K$ neighbors around $x^{0}$, and denote $\mathbb{X} = \{x^{0},x^{1}, x^{2}..., x^{K}\}$, where $\mathbb{X}$ means the set consisting of  $x^{0}$ and its $K$ neighbors.
Based on $\mathbb{X}$, $\{s^{0}, s^{1}, s^{2}..., s^{K}\}$, are derived as illustrated in Fig.~\ref{fig:method}.
Then, the derived score for voting, $s^{vote}$ is calculated as below:
\begin{align}
    &s^{vote} = \frac{1}{K+1} \sum_{k=0}^{K} s^{k} \label{eq:vote-score}
\end{align}
Based on  $s^{vote}$, the utterance will be accepted if $s^{vote} > \tau_{ASV}$, and vice versa.
Then the modified FAR and FRR for voting, $FAR_{vote}$ and $FRR_{vote}$, are derived:
\begin{align}
    &FAR_{vote} = \frac{\vert \{s^{vote}_{i} \geq \tau_{ASV} : i \in \mathbb{E}_{ntgt} \} \vert}{\vert \mathbb{E}_{ntgt} \vert} \label{eq:vote-far} \\
    &FRR_{vote} = \frac{\vert \{ s^{vote}_{i} < \tau_{ASV} : i \in \mathbb{E}_{tgt} \} \vert}{\vert \mathbb{E}_{tgt} \vert} \label{eq:vote-frr}
\end{align}
where $s^{vote}_{i}$ is the derived score for voting of the $i_{th}$ utterance.

\subsection{Why ``voting for the right answer" works}
\label{sec:vote_works}
It is hard to guarantee the well-trained ASV model is perfect.
So we can regard the ASV we derived is a slightly skewed mapping function from the data space to the label space, which makes some areas of the input space cannot be well covered by the ASV model.
The uncovered input space can be regarded as blind spots of ASV.
The malicious attackers deliberately try every attempt to find the blind spots and hallucinate the ASV model, and we think that adversarial samples are also blind spots of the ASV model.
Data augmentation is an intuitive way to enrich the model's distribution manifolds to cover the blind spots.
However, exhausting all the data augmentation strategies and making the model fill in all the manifolds of data distribution during training is extremely resource-consuming, not to mention that we can't model all the data augmentation strategies.
So we provide another alternative way to handle the blind spots: voting for the right answer.
In case that the testing sample falls into the blind spot, we propose to randomly sample $K$ samples within a Gaussian ball.
And then we let the neighbors of the testing sample vote for the right answer.
These areas of the blind spots are small in volume as shown in Table~\ref{tab:defense_results}, and when random sampling is conducted, the samples tend to jump out of the blind spots if the sampling variance, $\sigma$, is sufficiently large.
The sampled neighbours are with more probabilities to be in a health area rather than in a blind spot area.
Then after voting, the improved probability of being in a health area leads to improved probability of getting a 'normal' decision.

\subsection{Threat model and our countermeasures}
We divide the threat models according to the knowledge obtained by the attackers and followed by our countermeasures.

\textbf{Limited-knowledge attackers}: Limited knowledge attackers have the access to the internals of the target ASV model, including parameters and gradients, but are unaware of the defense method. The attackers elaborate the adversarial samples which are blind spots of ASV. In this scenario, our defense method can choose a suitable $\sigma$ and sample a large proportion of neighbours outside the blind spots, so that we have a large probability to get the right answer by voting.

\textbf{Sufficient-knowledge attackers}: Compared with limited-knowledge attackers, sufficient knowledge attackers even know the defense method.
They do gradient descent to modify the original sample and craft an adversarial counterpart to change the $s_{vote}$ in Eq.~\ref{eq:vote-score} to fool the ASV.
However, we conduct voting during inference and we don't know exactly the neighbors sampled for voting in advance, neither do the attackers. And it is impractical the attackers can exhaust all the possible neighbors for attack, so the exactly gradient for $s_{vote}$ during inference is hard to obtain. The randomness introduced by the voting procedure will hinder and obfuscate the gradient of $s_{vote}$ and make the gradient-based attacks ineffective. Our method attains more potential because we can even determine $\sigma$ and $K$ randomly.

\section{Experimental setup}
\label{sec:exp_setup}

\subsection{Data}

Voxceleb 1\&2 datasets \cite{nagrani2017voxceleb,chung2018voxceleb2} are used in our experiments.
We train models on the development set of VoxCeleb 1\&2. 
To test our ASV system, we use the trials provide in VoxCeleb1 test set, which contains 37,720 enrollment-testing pairs.
To conduct the experiment more rigorously and without loss of generality, we randomly select 27,720 trails as development set to evaluate the performance of our ASV system on genuine samples and determine the decision threshold in Eq.~\ref{eq:asv-threshold}, and the rest 10,000 trials are left as the evaluation set for generating adversarial samples and evaluating defense performance.

\subsection{ASV system setup}

We build our ASV systems following \cite{chung2020in}, using \textbf{Fast ResNet-34} model structures and angular margin (AM) loss.
Self-Attentive Pooling (\textbf{SAP}) \cite{zhu2018self} and Attentive Statistics Pooling (\textbf{ASP}) \cite{okabe2018attentive} are used to aggregate frame-level features into utterance-level representations.
\textbf{SAP} and \textbf{ASP} denote these two systems in the following description.
In our experiments, we set the hyperparameters of AM loss $scale$ as 30 and $margin$ as 0.1.
The models are trained for 50 epochs by the Adam optimizer with an initial learning rate of 0.01 decreasing by 10\% every 2 epochs.
During training, we use 2-second fixed length audio segments, extracted randomly from each utterance, and the batch size is 200. 
Spectrograms are extracted with a hamming window of width 25ms and shift 10ms, and 64-dimensional FBanks are used as the models' inputs. 
Our focus is countering the adversarial noise, rather than obtaining state-of-the-art ASV system. Therefore, no data augmentation and voice activity detection are conducted during training, and we use cosine distance for scoring.
Our model achieved 2.52\% equal error rate with ASP and 2.58\% with SAP in development set. 
We determine and fix the models' decision thresholds $\tau_{ASV}$ by Eq.~\ref{eq:asv-threshold}.

\subsection{BIM attack setup}
We generate adversarial samples by BIM. 
In our experiments, we fix the number of iterations $N$ as 5, set the step size $\alpha$ as $\epsilon$ divided by $N$, and use different settings of $\epsilon$: 1,5,10. 
The ASV systems' performance are reported in Table \ref{table:attack_results}. 
After attack, FAR and FRR increase drastically, which shows the effectiveness of BIM.
The larger $\epsilon$ is, the worse of the ASV performance, indicating more intense the attack is.

\begin{table}[h]
\centering
\caption{ASV performance for genuine and adversarial samples}
\begin{tabular}{cccccc}
\hline
\multicolumn{2}{c}{}       & No attack & $\epsilon=1$ & $\epsilon=5$ & $\epsilon=10$ \\ \hline
\multirow{2}{*}{\textbf{ASP}} & FAR & 2.24      & 18.38        & 71.83        & 89.38         \\
                     & FRR & 2.56      & 20.17        & 74.92        & 91.94         \\ \hline
\multirow{2}{*}{\textbf{SAP}} & FAR & 2.35      & 18.43        & 72.11        & 89.55         \\
                     & FRR & 2.23      & 20.21        & 74.33        & 91.7          \\ \hline
\end{tabular}
\label{table:attack_results}
\end{table}

\section{Experimental results}
\label{sec:exp_result}

\subsection{Defense for limited-knowledge attackers}

\begin{table*}[t]
\centering
\caption{Defense performance against Limited-knowledge attackers}
\resizebox{\textwidth}{17mm}{
\begin{tabular}{cccccccccccc}
\hline
\multicolumn{1}{l}{} & \multicolumn{1}{l}{} & \multicolumn{1}{l}{\multirow{2}{*}{No defense}} & \multicolumn{1}{l}{\multirow{2}{*}{Gaussian}} & \multicolumn{1}{l}{\multirow{2}{*}{Mean}} & \multicolumn{1}{l}{\multirow{2}{*}{Median}} & \multicolumn{6}{c}{Voting}                                                                  \\
                     &                      & \multicolumn{1}{l}{}                            & \multicolumn{1}{l}{}                          & \multicolumn{1}{l}{}                      & \multicolumn{1}{l}{}                        & $\sigma=1$    & $\sigma=15$ & $\sigma=30$   & $\sigma=60$    & $\sigma=90$ & $\sigma=120$   \\ \hline
\multirow{4}{*}{FAR} & No attack            & 2.24                                            & 1.88                                          & 2.69                                      & 0.98                                        & 2.25          & 2.32        & 2.16          & 2.14           & 1.84        & 1.61           \\
                     & $\epsilon=1$         & 18.38                                           & 11.47                                         & 12.47                                     & 1.85                                        & 16.36         & 27.33       & 2.89          & 2.42           & 2.13        & \textbf{1.83}  \\
                     & $\epsilon=5$         & 71.83                                           & 56.86                                         & 59.66                                     & 6.3                                         & 70.83         & 27.55       & 10.62         & 4.63           & 3.05        & \textbf{2.54}  \\
                     & $\epsilon=10$        & 89.38                                           & 80.7                                          & 82.89                                     & 12.35                                       & 89.13         & 69.2        & 29.66         & 9.43           & 5.32        & \textbf{3.6}   \\ \hline
\multirow{4}{*}{FRR} & No attack            & 2.56                                            & 3.04                                          & 2.05                                      & 19.55                                       & 2.27 & 2.82        & 3.32          & 4.95           & 6.06        & 8.12           \\
                     & $\epsilon=1$         & 20.17                                           & 17.38                                         & 14.77                                     & 24.24                                       & 16.59         & 4.9         & \textbf{4.53} & 5.33           & 6.98        & 8.76           \\
                     & $\epsilon=5$         & 74.92                                           & 62.22                                         & 65.47                                     & 40.8                                        & 72.79         & 33.56       & 16.78         & \textbf{10.74} & 10.81       & 12.18          \\
                     & $\epsilon=10$        & 91.94                                           & 87.58                                         & 87.13                                     & 55.42                                       & 91.1          & 73.74       & 42.52         & 21.77          & 17.05       & \textbf{16.67} \\ \hline
\end{tabular}
}
\label{tab:defense_results}
\end{table*}

In this scenario, the attackers can access the ASV model, but do not know the defense method.
There is few baseline for reference and in order to make a clear comparison, we follow our previous work \cite{wu2021adversarialasv} which applies hand-crafted filters including Gaussian filter, mean filter and median filter for adversarial defense, and set them as our baseline. 
We use different settings of the voting variance $\sigma$, randomly sample 50 neighbors around the testing audio, and let the neighbors vote for whether the utterance should be accepted or not.
The performance with \textbf{SAP} exhibits a similar trend as with \textbf{ASP}, and due to limited space, we only report the results of \textbf{ASP}.
In Table~\ref{tab:defense_results}, we evaluate the FAR and FRR of different defense methods, and note that the FAR and FRR for voting is calculated as Eq.~\ref{eq:vote-far} and Eq.~\ref{eq:vote-frr}.
We evaluate both the negative effects of the defense methods for genuine samples (rows labeled by No attack), 
and the positive effects on adversarial samples in different settings of $\epsilon$.
The observations and analysis are concluded as follows:
(1) Gaussian, mean and the voting method with different $\sigma$ slightly affect the FAR and FRR of genuine samples, while median filter enlarges the FRR from 2.56\% to 19.55\%, indicating that median filter highly distorts the speech signal.
(2) The three filter-based models help alleviate the adversarial attacks, but the defense performance of them is greatly outperformed by the voting-based defense.
What's more, as $\epsilon$ become large, the performances of the three filter-based methods are significantly degraded, while the voting based method still decrease FAR and FRR to a great deal.
(3) Note that when $\sigma$ is small, the voting strategy can't alleviate the adversarial noise, as the neighbors are still with large probability in the blind spot as claimed in Section~\ref{sec:vote_works}.
And when $\sigma$ increases, the FAR and FRR drop drastically, as a large proportion of neighbors are outside the blind spot and they will vote for the right answer.

\begin{figure}[h]
  \centering
  \centerline{\includegraphics[width=6cm]{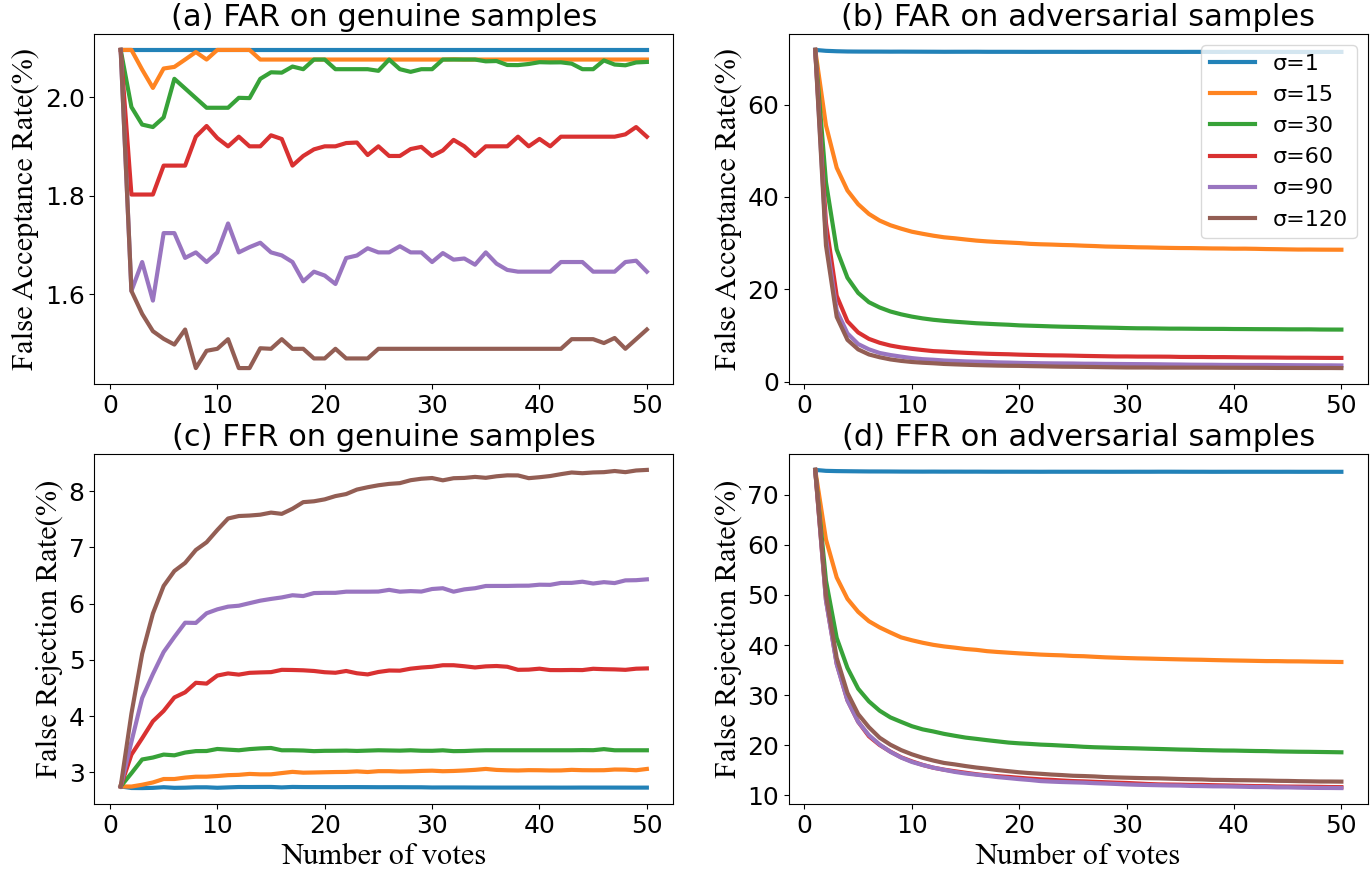}}
  \caption{Defense performance in different number of votes}
  \label{fig:num_votes}
\end{figure}

Fig.~\ref{fig:num_votes} shows FAR and FRR on genuine samples (a and c) and adversarial samples (b and d) with different number of votes, $K$, given $\epsilon = 5$.
We observe that:
(1) When the number of votes increases, the FAR and FRR for adversarial samples drops drastically.
And when $K$ is around 10, the FAR and FRR for adversarial samples become saturated, which means we can use about 10 votes to achieve effective defense performance.
(2) Fig.~\ref{fig:num_votes}.c shows as $K$ increases, the FRR on genuine samples increases, which means the voting strategy results in negative effect on genuine samples. However, this is just a mild negative effect outweighed by the performance improvement on adversarial samples.
We can choose a reasonable number $K$ to balance the trade-off between the negative effect on genuine samples and positive effect on adversarial samples, 
which take the system security and user experience into consideration according to the system requirements.

\subsection{Defense for sufficient-knowledge attackers}

In this subsection, we give a case study to evaluate the robustness of our method under a more extreme scenario, where attackers have sufficient knowledge in particular the defense method.
Sufficient knowledge BIM aims to attack not only
the original sample, but also its $K$ neighbors which are randomly sampled for voting.
Therefore, the new BIM against voting defense requires around $K+1$ times computing resource compared to the original BIM counterpart, however it's hard to accomplish by our computing resource when $K$ is too large \footnote[1]{This indicates when attackers have sufficient knowledge, the cost of attacking the system would increase.}.
So we just set the number of votes as 5 to illustrate a case study.
We fix $\epsilon$ as 5, attack iteration $N$ as 5, and $\sigma$ as 60.
For filter-based defense methods, the sufficient-knowledge attackers view the ASV system equipped with hand-crafted filters as an entire system, and directly get the gradient of such system and attack it.
Due to limited space, we only show the results of Gaussian filter, and the other filters are with the same trends.
Table~\ref{tab:perfect} reports the results.
Note that the performance of limited knowledge attackers in Table~\ref{tab:perfect} is worse than Table~\ref{tab:defense_results} because the numbers of votes are different.
We observe that:
(1) The voting based method achieves better defense performance than Gaussian filter.
(2) Though the attackers know the voting defense and use more computing resource, they still can't do attack effectively, as the FAR and FRR only increase slightly.

\begin{table}[ht]
\setlength\tabcolsep{4pt}
\centering
\caption{Defense results for sufficient-knowledge attackers }
\begin{tabular}{cccccc}
\hline
\multirow{2}{*}{} & \multirow{2}{*}{No defense} & \multicolumn{2}{c}{Gaussian} & \multicolumn{2}{c}{Voting} \\
                  &                             & limited       & sufficient      & limited      & sufficient     \\ \hline
FAR(\%)               & 71.83                       & 56.86         & 60.13        & 10.66        & 13.29       \\ \hline
FRR(\%)               & 74.92                       & 62.22         & 64.53        & 24.68        & 27.75        \\ \hline
\end{tabular}
\label{tab:perfect}
\end{table}

\begin{figure}[ht]
  \centering
  \centerline{\includegraphics[width=6cm]{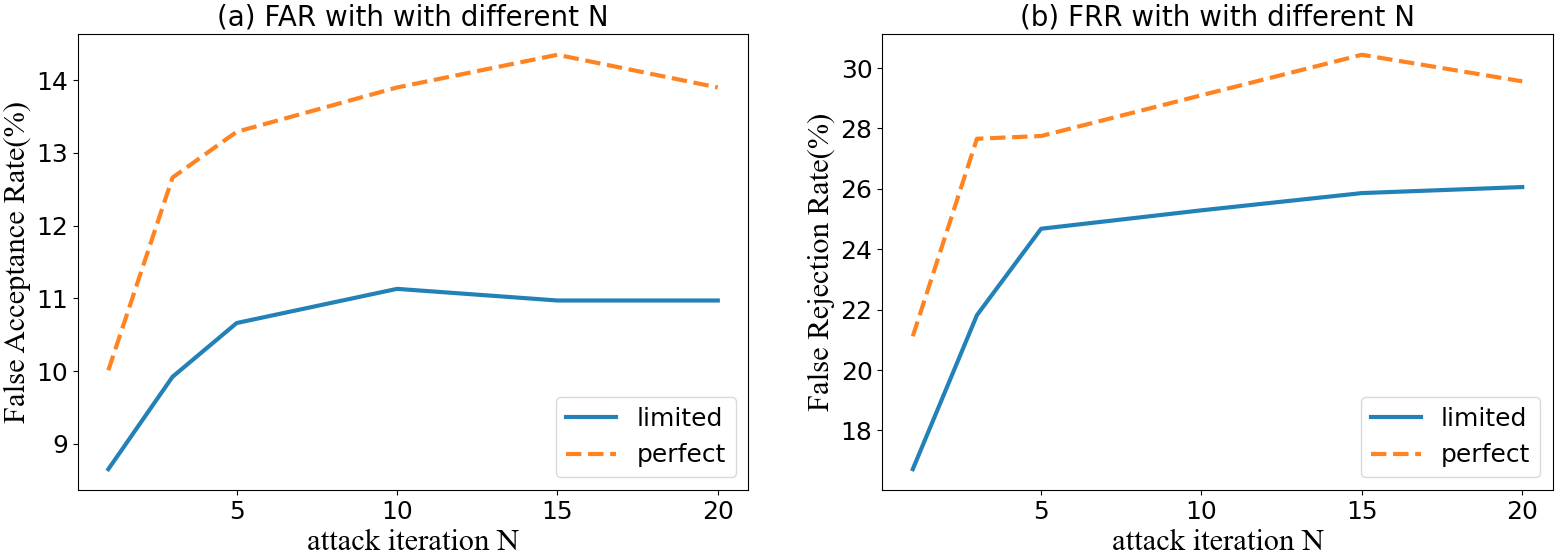}}
  \caption{Attack performance with different $N$}
  \label{fig:iter}
\end{figure}

Furthermore, Fig.~\ref{fig:iter} shows the attack results when the attack iteration $N$ increases, which means the attackers get more attack budgets by enlarging $N$.
However, as $N$ becomes larger than 10, the values of FAR and FRR become saturated.
This is another evidence to show the effectiveness of the voting based defense method, as it introduces randomness and obfuscates the gradient based attack algorithm.

\vspace{-10pt}

\section{Conclusion}


In this paper, we propose the idea of ``voting for the right answer" to prevent risky decisions of ASV in blind spot areas, by employing random sampling and voting.
Experimental results show our proposed method can both counter limited-knowledge attackers, and sufficient-knowledge attackers with more attack budgets.
We opened source the code 
\footnote[2]{\url{https://github.com/thuhcsi/adsv_voting}} to make our method a comparable baseline.
Future work will be dedicated to investigate other sample strategies e.g., sampling within the phone subspace, rather than the blind Gaussian to improve the defense performance of the voting strategy.
We are not familiar with adaptive attacks \cite{tramer2020adaptive} when writing this paper. We will refer to  \cite{yang2020characterizing} and dedicate efforts to invest more powerful adaptive attacks in the future work.

\section{Acknowledge}
This work was done when H. Wu was a visiting student at Shenzhen International Graduate School, Tsinghua University. H. Wu is supported by Frontier Speech Technology Scholarship of National Taiwan University. Y. Zhang is supported by National Natural Science Foundation of China (NSFC) (62076144) and joint research fund of NSFC-RGC (Research Grant Council of Hong Kong) (61531166002, N$\_$CUHK404/15).





\end{document}